\documentclass[letterpaper, 10 pt, conference]{ieeeconf}  

\IEEEoverridecommandlockouts                              
\usepackage[bookmarks=true]{hyperref}
\usepackage{tabularx}
\usepackage{graphicx}
\usepackage{epsfig}
\usepackage{subfig}
\usepackage{cite}
\usepackage{float}
\usepackage{ifthen}
\usepackage{xcolor}
\usepackage{ntheorem}
\usepackage{xcolor}
\theoremseparator{:}
\usepackage{soul}
\usepackage{colortbl}
\usepackage{booktabs}
\let\labelindent\relax
\usepackage{enumitem}

\usepackage[font=footnotesize]{caption}
\setlist[itemize]{leftmargin=*}
\usepackage[flushleft]{threeparttable}
\usepackage{enumerate}
\usepackage{algorithm}
\usepackage{algpseudocode}
\usepackage{cuted}
\usepackage{amsmath}
\usepackage{flushend} 

\newif\ifanonymous
\anonymousfalse   

\ifanonymous
    
\else
    
\fi

\newcommand{\fig}[1]{Fig.~\ref{#1}}

\def\eg{\emph{e.g., }} 
\def\ie{\emph{i.e., }} 

\begin{document}

\title{\LARGE \bf Understanding Whole-Body Robot Teleoperation Strategies Under Diverse Task Objectives and Constraints}

\ifanonymous
\author{Anonymous Authors}
\else
\author{Tsung-Chi Lin$^{1}$, Juo-Tung Chen$^{2}$, and Chien-Ming Huang$^{3}$
\thanks{$^{1}$Department of Computer Science, New Jersey Institute of Technology, Newark, NJ, USA.
{\tt\small tsungchi.lin@njit.edu}}%
\thanks{$^{2}$Department of Mechanical Engineering, Johns Hopkins University, Baltimore, MD, USA.
{\tt\small jchen396@jhu.edu}}%
\thanks{$^{3}$Department of Computer Science, Johns Hopkins University, Baltimore, MD, USA.
{\tt\small chienming.huang@jhu.edu}}%
}
\fi

\bstctlcite{IEEEexample:BSTcontrol}

\maketitle

\begin{strip}
\begin{minipage}{\textwidth}\centering
\ifanonymous
\else
    \vspace{-37pt}
\fi
\includegraphics[width=\textwidth]{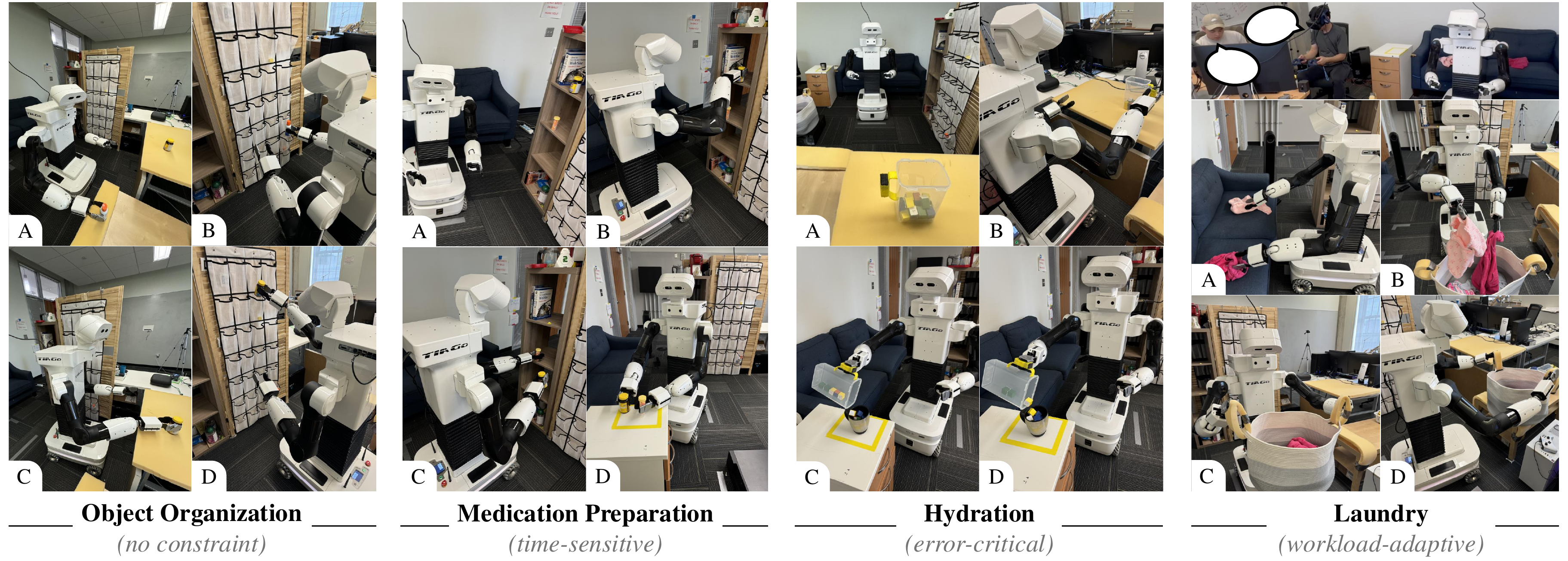} 
\captionof{figure}{To investigate the coordinated control strategies for robot teleoperation, users control the TIAGo robot to perform various daily home care tasks under different constraints. In the Object Organization scenario, the user organizes objects without limitations, exploring general control strategies in complex teleoperation. For Medication Preparation, the user handles medication bottles, focusing on timely delivery under time constraints. In the Hydration sequence, the user pours small blocks (i.e., simulating water) into a cup, where precision is crucial to avoid errors. Lastly, during the Laundry task, users control the robot to pick up and place clothing items into a basket and transfer it while conversing with the experimenter, simulating common daily distractions.}
\label{fig:tasks}
\end{minipage}
\end{strip}


\begin{abstract}

This work investigates the control strategies of complex whole-body robot teleoperation that coordinate active perception, bimanual manipulation, and navigation. We developed a hybrid control framework, combining the \textit{free-form} and \textit{constrained} control, for the whole-body teleoperation of the TIAGo mobile manipulator. We conducted a user study to explore people's control strategies under different task constraints such as limited time and low tolerance of errors. Our results highlight the effective use of coordinated control in improving task efficiency and reducing the risk of reaching individual joint limits.
We discuss our results and their implications for designing future whole-body robot teleoperation systems.

\end{abstract}

\section{Introduction}\label{sec:intro}

Teleoperated robots significantly augment human capabilities in scenarios where direct human presence is challenging (\eg deep-sea exploration~\cite{li2023bioinspired}) or complete robot autonomy is not reliable (\eg home care~\cite{toedtheide2023force}), allowing for the safe robot-assisted execution of complex tasks across diverse domains~\cite{darvish2023teleoperation}. 
In support of effective robot teleoperation, previous research has investigated various control interfaces for direct teleoperation of complex robots (\eg humanoid robots and mobile manipulators with high degrees of freedom).
For example, early research focused on motion tracking systems, which allow operators to control robots through their own movements using wearable sensors~\cite{zhou2019iot}, or vision-based systems~\cite{rakita2017motion}. 
Recent research explored virtual reality-based methods to control mobile manipulators~\cite{bejczy2020mixed} or humanoid robots~\cite{wonsick2021human}, offering not only greater portability and cost efficiency but also creating more immersive remote environments, which can be easily integrated with haptic~\cite{lin2022comparison} or augmented reality~\cite{livatino2021intuitive} features. 

Despite the ease of use of these control interfaces~\cite{lin2022intuitive}, coordinating the robot's multi-functional capabilities remains a difficult task. 
Prior works have investigated strategies to enhance loco-manipulation through shared autonomy~\cite{wu2019teleoperation,seo2023deep}, where the robot assists the human operator by automating certain functions while leaving critical decision-making to the human. Although these strategies have shown promise, the optimal approach for coordinated control---where perception, manipulation, and navigation are seamlessly integrated---remains unclear. 
Furthermore, it is not well understood how control strategies may change depending on task constraints (\eg limited time and low tolerance of errors) and requirements (\eg handling concurrent tasks).

This paper investigates human control strategies for teleoperating a complex robot with multi-functional capabilities; specifically, our investigation uses the TIAGo OMNI++ mobile manipulator robot with a controllable head, torso, and base, as well as two seven degree-of-freedom arms.
We implemented a hybrid control framework for whole-body robot teleoperation using a virtual reality (VR) system (\ie HTC Vive Pro 2) to leverage the strengths of both free-form (\ie multi-axis pose control) and constrained (\ie dual-axis touchpad control) methods (Fig.~\ref{fig:interface}). The robot’s head and dual arms are controlled via the pose of a VR headset and handheld controllers, while the torso and base are managed using the touchpads on both handheld controllers. A video stream from the head-mounted camera, integrated with the robot's operational state, is sent to the VR headset for real-time active remote perception. 

To understand people's control strategies, we conducted a user study in which participants performed four in-home care tasks with diverse objectives and constraints (\ie no constraint, time-sensitive, error-critical, and workload-adaptive) after completing a curriculum-based training designed to help novice users acquire basic competency in teleoperating the robot (Fig.~\ref{fig:tasks}).
Our results reveal several coordinated control strategies used and how they were adapted to different task constraints. We discuss the implications of these findings for control interface design, action support, and adaptive robot (shared) autonomy in teleoperated human-robot interactions. 
The key contributions are: 
1) a modularized hybrid manual control framework for complex whole-body robot teleoperation; 
2) a curriculum-based training for comprehensive teleoperation skill acquisition; and 
3) design implications for future human-centered system aimed to support whole-body robot teleoperation.

\section{Background and Related Work}\label{sec:RelatedWork}

\subsection{Whole-Body Teleoperation Interface}
As robots become more physically capable~\cite{fu2024mobile}, the need for control interfaces that can handle whole-body coordination—including simultaneous manipulation, locomotion, and sensor alignment—has become increasingly important. Whole-body teleoperation introduces unique interaction demands, requiring users to synchronize perception with action across multiple control dimensions in real time. To this end, the evolution of teleoperation interfaces for whole-body robot control has seen significant advancements, particularly in the development of constrained, free-form, and hybrid control approaches. 

\textit{Constrained control} interfaces simplify operation by limiting the robot's degrees of freedom (DoF) and restricting its movements to predefined paths or regions, allowing for easier control in specific tasks like manipulation~\cite{dall2025towards} or navigation~\cite{raheema2024autonomous}. Although these approaches excel in simplicity, they often limit the robot's full capabilities, particularly in complex tasks requiring high maneuverability~\cite{lin2022intuitive}. In contrast, \textit{free-form control} interfaces offer complete flexibility, enabling users to directly control the robot’s various components; prior research has focused on motion tracking systems that allow operators to control robots with their own movements via wearable sensors~\cite{ze2025twist}, exosuits~\cite{dafarra2024icub3}, kinesthetic interfaces~\cite{fu2024mobile}, and vision-based systems~\cite{he2024learning}. But despite their flexibility, these systems often come with the trade-off of increased cognitive load and complexity, as users must manage multiple control aspects simultaneously; having to prevent unintentional movement can also be a barrier, making it difficult for non-expert users to operate such systems effectively~\cite{lin2024perception}. To this end, \textit{hybrid control} approaches have been developed by combining the strengths of constrained and free-form methods. In particular, immersive virtual- and mixed-reality teleoperation systems have demonstrated how spatially aligned visual feedback and embodied input can support intuitive control while reducing cognitive translation between the operator and the robot. Prior work has explored VR-based teleoperation for manipulation and navigation tasks, examining user performance, workload, and sensitivity to latency and interface design in immersive settings~\cite{naceri2021vicarios}. More recent systems extend these ideas to whole-body platforms, introducing hybrid input strategies that integrate body navigation with head-and-arm control for mobile manipulators~\cite{park2023whole} and humanoid robots~\cite{penco2024mixed}.

However, it remains unclear how teleoperation strategies adapt to the complexity of real-world tasks with varying objectives and constraints. Without understanding the interplay between human factors and task requirements, it is difficult to design teleoperation systems that provide optimal support tailored to different user groups and everyday tasks.

\subsection{Multi-Component Coordination in Complex Teleoperation}
Whole-body teleoperation tasks require human operators to coordinate multiple robotic components—including perception (head), manipulation (arms), locomotion (base), and vertical positioning (torso)—in real time. This coordination is especially critical for mobile manipulators and humanoid systems, where performance often depends on precise synchronization across spatial and functional domains~\cite{honerkamp2025whole}. Unlike single-arm or base-only teleoperation, whole-body tasks present increased cognitive and physical challenges due to their inherently high degrees of freedom and interdependent control actions~\cite{gholami2022quantitative}.

A growing body of research has highlighted the complexity of this problem by examining how users coordinate multiple robot subsystems during remote interaction. One prominent challenge involves bimanual coordination, where users must simultaneously manage two robotic arms for tasks such as lifting, folding, or transferring objects. Recent work has investigated shared-autonomy frameworks for intuitive bimanual telemanipulation, enabling users to perform coordinated dual-arm actions with reduced control effort by blending user intent with robot policy priors~\cite{laghi2018shared}. Similarly, single-master systems have been proposed to allow effective bimanual control using one interface, reducing redundancy and improving task efficiency~\cite{sun2020single}.

Beyond the arms, coordinating the torso and arms is essential when users must extend the robot’s vertical reach or avoid kinematic singularities. Effective torso-arm coordination is particularly important in manipulation tasks that span shelves, tables, or uneven surfaces. For example, recent work on mobile humanoid control demonstrated how integrating torso adjustments with arm movements can facilitate reaching and manipulation without sacrificing stability or control intuitiveness~\cite{boguslavskii2025human}. Without such coordination, users must frequently interrupt manipulation to adjust posture manually, leading to inefficiencies and increased mental workload.

A third critical axis of coordination lies in manipulation-locomotion coupling, where users must synchronize base movement with arm actions to approach, interact with, or reposition objects. This coordination is especially relevant in cluttered or dynamic environments. Prior work has proposed whole-body teleoperation approaches that allow users to control navigation and manipulation simultaneously without requiring additional interface complexity~\cite{honerkamp2025whole}. These systems show that tightly coupled control of arm and base can reduce task time and physical effort—but only when the interface effectively supports fluid transitions and alignment between components.

These studies illustrate that whole-body teleoperation is not simply about exposing control over individual components, but about enabling operators to effectively manage the interdependencies between them. However, much of the current literature evaluates component coordination in isolation or under idealized tasks. How users adapt coordination strategies in response to task constraints, such as time pressure, accuracy requirements, or workload, remains underexplored. Understanding these dynamics is key to designing teleoperation systems that can adaptively support the user's coordination behavior rather than demanding constant micromanagement.

\section{Robot Teleoperation System}\label{sec:proposed}

Below, we describe the implementation of a hybrid control framework for complex teleoperation and the design of the graphical user interface for remote perception.

\begin{figure}[t]
    \centering
    \includegraphics[width=1\linewidth]{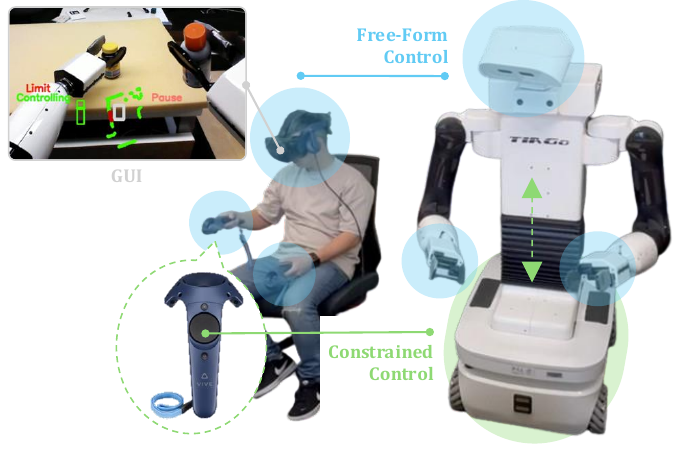}
    \caption{The hybrid control teleoperation interface captures human head and hand movements to control the robot's head and dual arms, while the touchpads on both handheld controllers control the robot's torso and base. The GUI integrates real-time camera streams and robot status.}
    \label{fig:interface}
    \vspace{-2ex}
\end{figure} 

\subsection{Hybrid Control Framework for Whole-body Teleoperation}

We implemented a modular hybrid control framework on a physical testbed using the TIAGo OMNI++ mobile manipulator, which integrates key robotic capabilities—including perception, manipulation, and locomotion—with an adjustable torso. The framework is designed to be modular and adaptable across a broad range of robot platforms (from robotic arms to humanoid robots) and multi-component configurations with varying input modalities.

Our system uses the HTC Vive Pro 2 virtual reality headset and controllers, which provide 6-DoF pose tracking, button states, and dual-axis touchpad inputs. The hybrid control approach combines \textit{free-form control} for the robot's head and arms with \textit{constrained control} for the torso and base (\fig{fig:interface}). Utilizing free-form control proved to be more intuitive for users new to robot teleoperation when handling complex features, enhancing egocentric perception~\cite{lin2023perception} and efficient manipulation~\cite{lin2022intuitive}. In contrast, constrained control via touchpads enables precise and stable movements, which are critical for tasks requiring accurate positioning and alignment, such as navigating tight environments or approaching objects. 

\begin{algorithm}[t]
\caption{Whole-body Teleoperation Control Loop}
\label{alg:hybrid_control}
\footnotesize
\algrenewcommand{\algorithmicindent}{1em}
\begin{algorithmic}[1]
\State Initialize VR system and control mappings
\While{system is active}
    \State Read head pose $(\mathbf{p}_h,\mathbf{R}_h)$
    \State Read controller poses $(\mathbf{p}_c^{(i)},\mathbf{R}_c^{(i)})$,
    \Statex \hspace{\algorithmicindent} for $i\in\{L,R\}$
    \State Read touchpad inputs and button states

    \Statex \Comment{\textbf{Head Control (Free-form)}}
    \State $\mathbf{R}_{\mathrm{head}} \gets
        \mathcal{L}(\mathbf{R}_h\operatorname{Rot}_x
        (\theta_{\mathrm{offset}}))$
    \State Send $\mathbf{R}_{\mathrm{head}}$ to robot head

    \Statex \Comment{\textbf{Arm Control (Free-form)}}
    \ForAll{$i\in\{L,R\}$}
        \If{grip button $i$ is pressed}
            \If{pause toggled}
                \State Reset control origin for arm $i$
            \EndIf
            \State $\mathbf{x}_{ee}^{(i)} \gets$ relative pose of controller $i$
            \State $\mathcal{J}^{(i)} \gets \mathcal{IK}(\mathbf{x}_{ee}^{(i)})$
            \If{arm $i$ extension $>\delta_{\mathrm{arm}}$}
                \State Display user warning
            \EndIf
            \State Send $\mathcal{J}^{(i)}$ to robot arm $i$
            \If{trigger $i$ is pressed}
                \State Close gripper $i$
            \Else
                \State Open gripper $i$
            \EndIf
        \EndIf
    \EndFor

    \Statex \Comment{\textbf{Torso Control (Constrained)}}
    \If{right touchpad: up}
        \State $T_z \gets T_z+\Delta z$
    \ElsIf{right touchpad: down}
        \State $T_z \gets T_z-\Delta z$
    \EndIf
    \State $T_z \gets \min(T_{\max},\max(T_{\min},T_z))$

    \Statex \Comment{\textbf{Base Control (Constrained)}}
    \State $\mathbf{v}_b \gets \mathcal{M}_{\mathrm{pad}}(u_{tp}^{L})$
    \State $\omega_b \gets \mathcal{M}_{\mathrm{pad}}(u_{tp}^{R})$

    \Statex \Comment{\textbf{Safety-Aware Damping}}
    \If{arm extension $>\delta_{\mathrm{arm}}$ \textbf{or}
        $T_z>\delta_{\mathrm{torso}}$}
        \State $\mathbf{v}_b \gets \alpha\mathbf{v}_b$,\quad
               $\omega_b \gets \alpha\omega_b$
    \EndIf
    \State Send $(\mathbf{v}_b,\omega_b)$ to robot base
    \State Update GUI
\EndWhile
\end{algorithmic}
\end{algorithm}

\noindent
\textbf{Free-form Control for Head and Arms:}
The user's head orientation is mapped to the robot’s pan-tilt head, with a downward offset of $\theta_{\text{offset}}$ degrees and smoothing via a low-pass filter to reduce fatigue and jitter. This adjustment was made because the focused workspace is typically at a similar height to the arms. Without the offset, users would need to actively maintain a bent neck position, significantly increasing physical fatigue during prolonged tasks. Additionally, setting a head mapping offset makes it easier to visualize the arms in the graphical user interface, enhancing awareness of arm location and potential collisions.

Each robotic arm is controlled by mapping the handheld controller's pose to the desired end-effector pose, solved using the TRAC-IK solver~\cite{beeson2015trac}. Users could reset the arm configuration to the starting pose (see~\fig{fig:interface}) at any time by pressing the menu button; activate or pause arm control with the grip button; and control the gripper with the trigger button on each handheld controller. Arm extension is continuously monitored and alerts the user when a threshold $\delta_{\text{arm}}$ is exceeded. The gripper is actuated by a binary trigger input.

\noindent
\textbf{Constrained Control for Torso and Base:}
The robot’s torso is incrementally adjusted in height via up/down presses on the right touchpad, moving by step size $\Delta z$. The mobile base is controlled translationally using the left touchpad and rotationally using left/right presses on the right touchpad. To improve safety, a damping mechanism reduces base speed when arm extension or torso height exceeds configurable thresholds ($\delta_{\text{arm}}, \delta_{\text{torso}}$). This ensures stable operation when the robot's center of mass is raised or the arms are near their limits. To formalize the system behavior, we present the full control loop in Algorithm~\ref{alg:hybrid_control}, and provide a summary of control mappings and parameters in Table~\ref{tab:control_mapping}. 

\begin{table}[t]
\caption{Summary of Control Mappings and System Parameters}
\label{tab:control_mapping}
\centering
\begin{tabular}{@{}lll@{}}
\toprule
\textbf{Component} & \textbf{Control Input} & \textbf{Parameters} \\
\midrule
Head            & Headset rotation          & $\theta_{\text{offset}} = 30^\circ$ \\
Arms            & Controller 6-DoF pose     & $\delta_{\text{arm}} = 0.7$ m \\
Gripper         & Trigger button (binary)   & -- \\
Torso height    & Touchpad up/down          & $\Delta z = 0.05$ m \\
Base translation& Touchpad (left)           & $\alpha = 0.5$ damping factor \\
Base rotation   & Touchpad (right, L/R)     & $\delta_{\text{torso}} = 0.5$ m \\
\bottomrule
\end{tabular}
\vspace{-2ex}
\end{table}

\subsection{Graphical User Interface (GUI)}

The GUI integrates real-time video feedback from an RGB camera built into the TIAGo's head. It overlays the robot's operational states, streaming them to the HTC Vive headset via the User Datagram Protocol (UDP). These operational states include arm status, torso height indicators, and obstacle visualization derived from object detection using a LiDAR sensor on the mobile base (\fig{fig:interface}).

\noindent
\textbf{Visualization of Proximal Obstacles:} 
To provide real-time feedback on the robot's movements and obstacle awareness, we created a dynamic mini-map that continuously updates to show obstacles within the robot's workspace, with a color change to red indicating too close to an obstacle (less than 0.1 meters). This approach provides a real-time reflection of the robot's movements and obstacle awareness.

\noindent
\textbf{Visualization of Torso Elevation and Arm State:} 
In addition to the obstacle visualization feature mentioned above, our GUI also includes a torso height indicator, which provides real-time feedback on the robot's vertical position, crucial for tasks requiring different elevation levels. Additionally, the status of both arms is prominently displayed, showing whether each arm is actively being controlled or in a paused state. The interface also alerts the user when the robot hits its operational limits, ensuring immediate awareness and response to any constraints. 

\section{Experiment}\label{sec:exp}

We conducted a user study to investigate control strategies for teleoperating a complex robot capable of multi-functional capabilities using our hybrid framework. In our study, we used home care tasks as they represent diverse constraints and objectives. A deeper understanding of people's control strategies will shed light on how to avoid unnecessary duplication of control inputs, suboptimal performance outcomes, and potentially inappropriate teleoperation assistance. 

\subsection{Curriculum-Based Training}\label{sec:training}

\begin{figure*}[t]
    \begin{center}
    \vbox{ 
    \includegraphics[width=\textwidth]{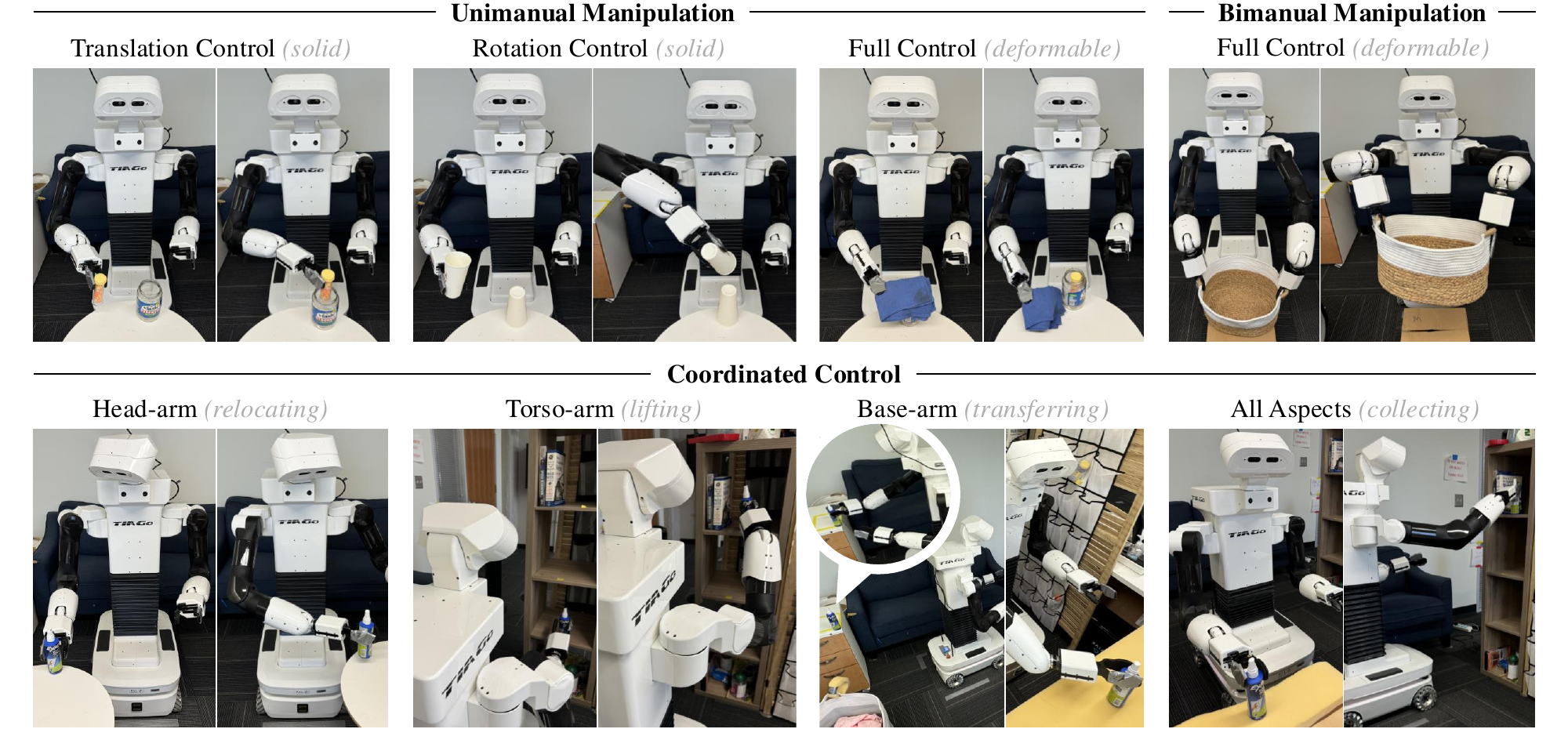} 
    }
    \caption{The curriculum-based training for manipulation and coordination in complex robot teleoperation tasks. The training is structured into two main categories: manipulation and coordination. In the manipulation category, tasks are divided into unimanual, where the user performs tasks using a single arm, including handling solid objects (e.g., a tube container or cup) and deformable objects (e.g., a soft rag), and bimanual, where the user uses both arms for tasks such as lifting a large object and collecting multiple items. In the coordination category, tasks involve the combined use of different robot components, including head-arm coordination, torso-arm coordination, base-arm coordination, and head-torso-base-arm coordination, to perform tasks.}
    \label{fig:training}
    \vspace{-2ex}
    \end{center}
\end{figure*}

The variance in users' prior experiences with gaming, virtual reality, and robotics might significantly impact their robot teleoperation performance. 
Thus, to ensure a relatively uniform proficiency level~\cite{lin2024targeted}, we designed a curriculum-based training protocol (see~\fig{fig:training}) to comprehensively guide novice users in acquiring the necessary skills. This training approach has been proven effective for gaining essential knowledge in various remote robot control applications (\eg space robots~\cite{liu2013predicting}, surgical robots~\cite{dulan2012developing}, and drones~\cite{zhou2019impact}). The curriculum is divided into two categories: basic manipulation skills and advanced coordinated control. Each category contains multiple training modules, with each module building on the previous one, aiming for a gradual increase in difficulty and complexity.

\noindent
\textbf{Basic Manipulation Skills:} 
The training begins with unimanual manipulation and progresses to bimanual manipulation, allowing users to acquire the necessary skills to interact with diverse targets (Fig.~\ref{fig:training}, top). During this training, the robot remains in a fixed location with a consistent torso height for all participants. The robot's head remains controllable by users, as the tasks are primarily within a small workspace that does not require large-scale head movement.

\begin{itemize}

    \item \textbf{\textit{Unimanual Manipulation.}} Unimanual manipulation training includes modules for various robot end-effector controls to interact with both solid and deformable objects. The \textit{translation} control module tasks users with picking up a small solid container and dropping it into a jar. The \textit{rotation} control module asks users to flip a held cup and stack it onto another cup placed on the table. Finally, the \textit{full} control module requires users to grab a rag from atop a jar and place it on the table.
    
    \item \textbf{\textit{Bimanual Manipulation.}} The bimanual manipulation training requires users to perform a large basket handover task. Users must lift the basket to a specific height and hand it over to an experimenter standing in front of the robot. This task necessitates full control of both robot arms and symmetric bimanual control.
    
\end{itemize}

\noindent
\textbf{Advanced Coordinated Control:}
Coordinated control training is essential to enable robots to perform complex tasks that require synchronized movement of multiple components. This training prepares participants for complex teleoperation where precise and coordinated actions are crucial for task completion (Fig.~\ref{fig:training}, bottom). 

\begin{itemize}

    \item \textbf{\textit{Head-arm Coordination.}} This training involves tasks where participants need to relocate an object by teleoperating significant robot arm movement while controlling the robot's head to maintain visual contact with the object so as being able to adjust arm movements accordingly. 
    
    \item \textbf{\textit{Torso-arm Coordination.}} The training focuses on object lifting tasks where participants need to control the robot, coordinating its torso and arm movement, to grasp an item from a lower shelf and place it on a higher level. 
    This training requires adjusting the torso height to accommodate the enlarged vertical workspace. 

    \item \textbf{\textit{Base-arm Coordination.}} This training involves searching for an object within the workspace and transferring it to a new location through a relatively narrow path. 
    The training task requires the teleoperated robot to perform large movements for search-and-transport and small adjustments to avoid collisions with obstacles. 
    Effective base-arm coordination ensures that the robot can maneuver through a constrained space for object manipulation.

    \item \textbf{\textit{Head-torso-base-arm Coordination.}} This training involves picking an object from the top of a lower table and placing it on a higher level of a shelf located at a distance. 
    The task requires participants to coordinate the robot's head, torso, arm, and base movements. Such coordination is essential for handling intricate real-world tasks that demand multi-component synchronization.
    
\end{itemize}

\subsection{In-home Care Tasks}

To ensure the practical relevance and generalizability of our findings, we designed four in-home care tasks with various constraints and objectives (see~\fig{fig:tasks}). Specifically, participants control the TIAGo robot in a cluttered environment to perform unstructured tasks without constraints, time-sensitive tasks, error-critical tasks, and multi-tasking tasks. Note that for tasks involving multiple targets, there are no restrictions on the interaction order or specific arm usage.

\noindent
\textbf{Object Organization \textit{(no constraint)}:}
Participants operate a robot to gather objects from two tables of varying heights and deposit them into predefined pockets on a wall-mounted organizer without constraints. This task allows us to probe baseline strategies for the control of a complex robot.

\noindent
\textbf{Medication Preparation \textit{(time-sensitive)}:} 
Participants are tasked with delivering two medication bottles placed on two different levels of a shelf to a marked region within a five-minute time limit. The control GUI includes a countdown timer to remind users of the remaining time. This task aims to examine people's control strategies under time pressure.

\noindent
\textbf{Hydration \textit{(error-critical)}:} 
In this task, participants control the robot to pick up a bottle containing blocks (simulated water) and pour them into a mug; precision is crucial to avoid water spills. This task seeks to understand control strategies for minimizing teleoperation errors.

\noindent
\textbf{Laundry \textit{(workload-adaptive)}:} 
This task involves participants controlling the robot to gather and drop clothing items into a basket, then move it to the top of a table. 
During the main control task, the experimenter engages the participant in a secondary task by chatting with them, discussing randomly chosen topics from ten general themes (\eg favorite travel destinations, hobbies, and interests, etc.). 
This scenario simulates real-world conditions where operators manage varying workloads and distractions (\eg answering phone calls).

\subsection{Procedure}
After obtaining informed consent, participants completed a demographic survey and assessed their spatial ability using the Purdue Spatial Visualization Rotation Test, which has been used in the context of teleoperation~\cite{menchaca2007influence}. This test consists of ten questions with a five-minute time limit. Next, the experimenter explained how to control the TIAGo robot using the HTC Vive VR headset and handheld controllers. Once participants indicated readiness, they began the curriculum-based training to acquire manipulation and coordinated control skills. During training, participants could repeat specific modules for additional practice, with this training phase lasting up to 30 minutes. Following training, participants were introduced to four in-home care tasks. After completing each task in random order, participants filled out the NASA-TLX and a questionnaire to provide feedback on their experience with the tasks.

\begin{figure}[b]
    \centering
    \includegraphics[width=\linewidth]{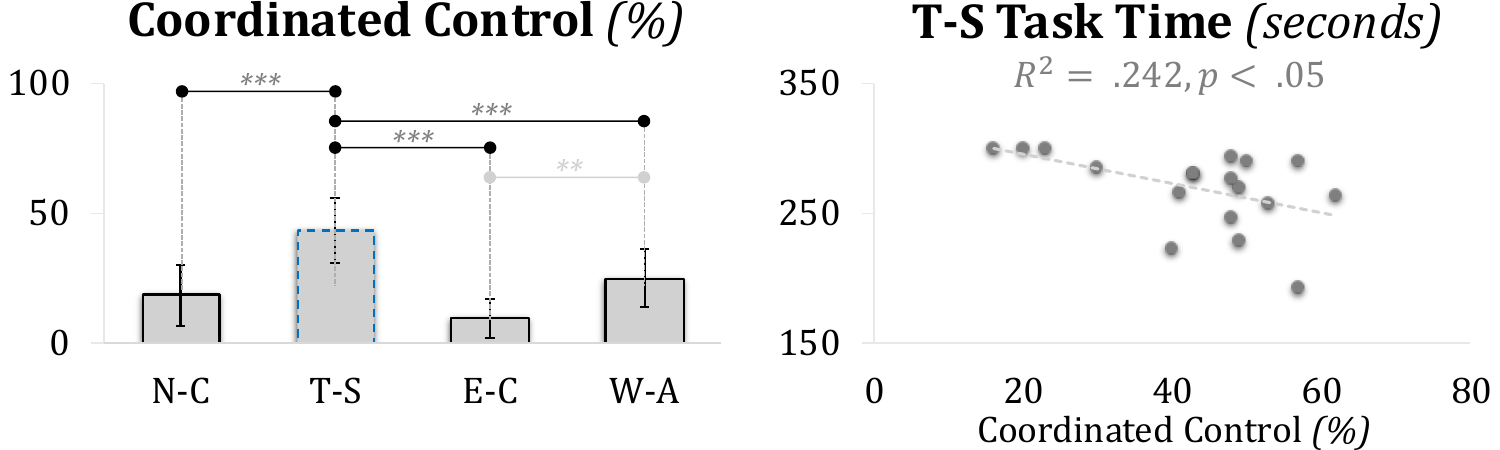}
    \caption{Coordinated control percentage across four tasks (left) and its correlation with task time in the time-sensitive task (right).}
    \label{fig:coordinated}
\end{figure} 

\subsection{Measures and Data Analysis}
We measured a set of control-, robot-, and task-related indices to objectively analyze participants' control strategies and their potential correlation with task performance. 

\noindent
\textbf{Control-related:} 
We recorded the activation time for each robot component (\ie head, torso, arms, and base) and the duration of the coordinated control (\ie head-arm, head-base, torso-arm, base-arm) for each task. Additionally, the users' control input speed was calculated from the recorded handheld controller's trajectories.

\noindent
\textbf{Robot-related:}
We measured how often the robot was near surrounding obstacles (\ie within a distance of less than 0.1 meters). We also recorded the duration for which the robot's arms and head reached their length and angle limits.

\noindent
\textbf{Task-related:}
We measured the time and the number of errors that occurred while performing each task. Errors included collision with or knocking down the target, spills of simulated water, and unsuccessful grasping and releasing. We extracted the mental demand from the NASA-TLX to understand the cognitive effort while performing each task.

We additionally used a questionnaire to gather participants' perceptions of the challenging aspects of teleoperating the robot. The questionnaire also inquired about which parts of the tasks or specific actions/motions they preferred to be handled by robot autonomy. 

We used a mixed regression model to assess the correlation between control-related and robot- or task-related measures. We analyzed all measures from each in-home care task as the within-participants variable using one-way repeated-measures analysis of variance (ANOVA) for all comparisons. All pairwise comparisons used Holm-Bonferroni correction to control for Type I error in multiple comparisons.

\subsection{Participants}

\begin{figure}[b]
    \centering
    \includegraphics[width=\linewidth]{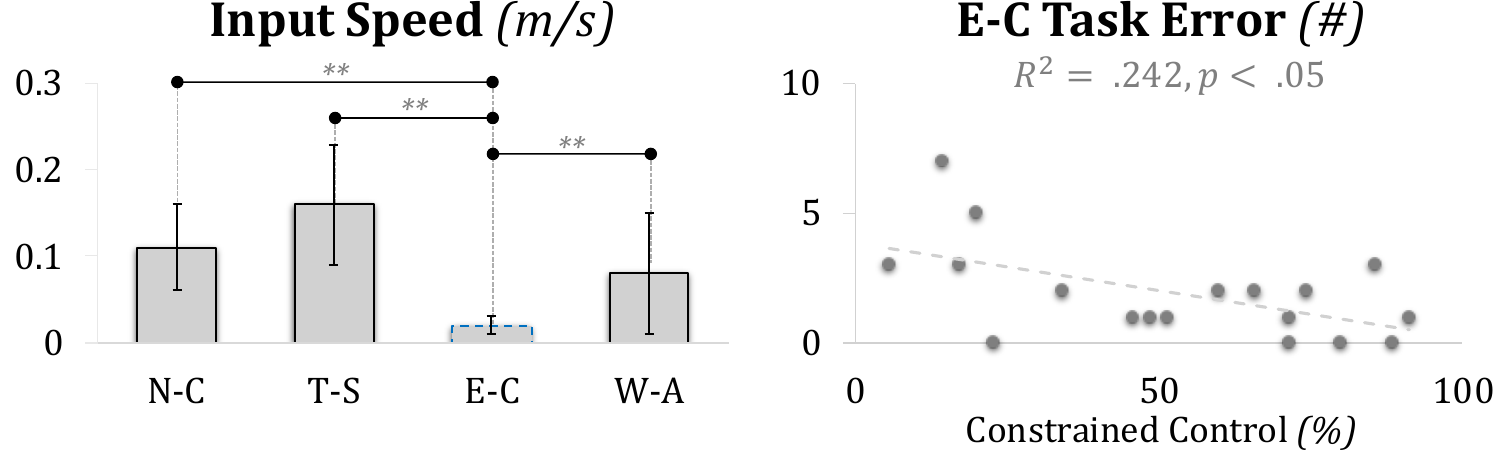}
    \caption{Input speed across four tasks (left) and the correlation between constrained torso/base control and error rate in the error-critical task (right).}
    \label{fig:precision}
\end{figure} 

We recruited 18 participants (8 males and 10 females) from the Johns Hopkins University Homewood campus, with ages ranging from 19 to 39 (\textit{M} = 26.72, \textit{SD} = 5.35). Participants reported limited prior experience with robots (\textit{M} = 1.92, \textit{SD} = 0.34) and diverse driving (\textit{M} = 2.27, \textit{SD} = 2.15) and gaming (\textit{M} = 1.83, \textit{SD} = 2.33) experiences, measured on a five-point scale with 5 being highly experienced. 
 
\section{Results and Discussion}\label{sec:res}

This section examines the impacts of task objectives and constraints on performance and control behaviors. We discuss their implications for the design of future robot teleoperation systems. In all figures, $p < .01$ and $p < .001$ are represented by two stars (**) and three stars (***). Task constraints are denoted as N-C (no constraint), T-S (time-sensitive), E-C (error-critical), and W-A (workload-adaptive).

\subsection{Task Efficiency}

We averaged the time spent on head-arm, torso-arm, and base-arm as coordinated control. The results show that users performed significantly more coordinated control in the time-sensitive task (\fig{fig:coordinated}, left) than in the other three tasks (all with $p < .001$ from post hoc comparisons). Correlation analyses further revealed that increased coordinated control significantly reduced ($p < .05$) task completion time (\fig{fig:coordinated}, right) and decreased the frequency of the robot reaching its arm length limit ($p < .05$). The analysis also revealed that the duration for hitting robot's head pan limit was significantly less ($p < .05$) with more head-base coordination control. 

\textit{User-centered Control ---} Our findings indicate that leveraging coordinated control allows for efficient task completion and reduces the risk of exceeding joint limits, thus avoiding redundant control actions. Building upon the loco-manipulation concept~\cite{wu2019teleoperation}, future teleoperation systems should integrate control frameworks that leverage a user's egocentric inputs (\eg head and hand motions) to manage not only the robot's head and manipulators but also other operational aspects, such as torso and base, for coordinated whole-body teleoperation. For example, when searching for an object, the robot base should rotate when the user reaches their head pan limit, aligning with their intent to adjust the viewing angle. Similarly, the torso and base should automatically adjust when the arms reach their limits to improve reachability in large workspaces. This user-centered approach reduces cognitive workload by minimizing the need to control multiple robot aspects while maintaining effective teleoperation. It also aligns with advanced VR/AR interfaces, like Apple Vision Pro, which leverage natural head and hand movements over traditional controllers.

\subsection{Task Precision}

\begin{figure}[t]
    \centering
    \includegraphics[width=\linewidth]{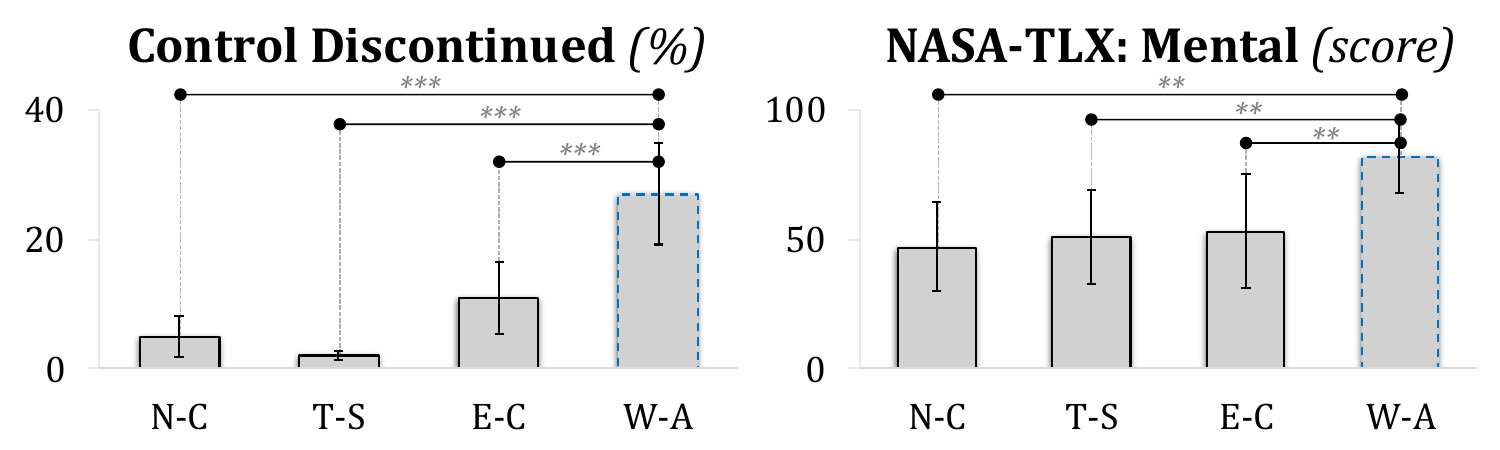}
    \caption{Percentage of control discontinuation (left) and mental demand from the NASA-TLX survey (right) across four tasks.}
    \label{fig:workload}
    \vspace{-2ex}
\end{figure} 

\begin{figure}[b]
    \centering
    \includegraphics[width=\linewidth]{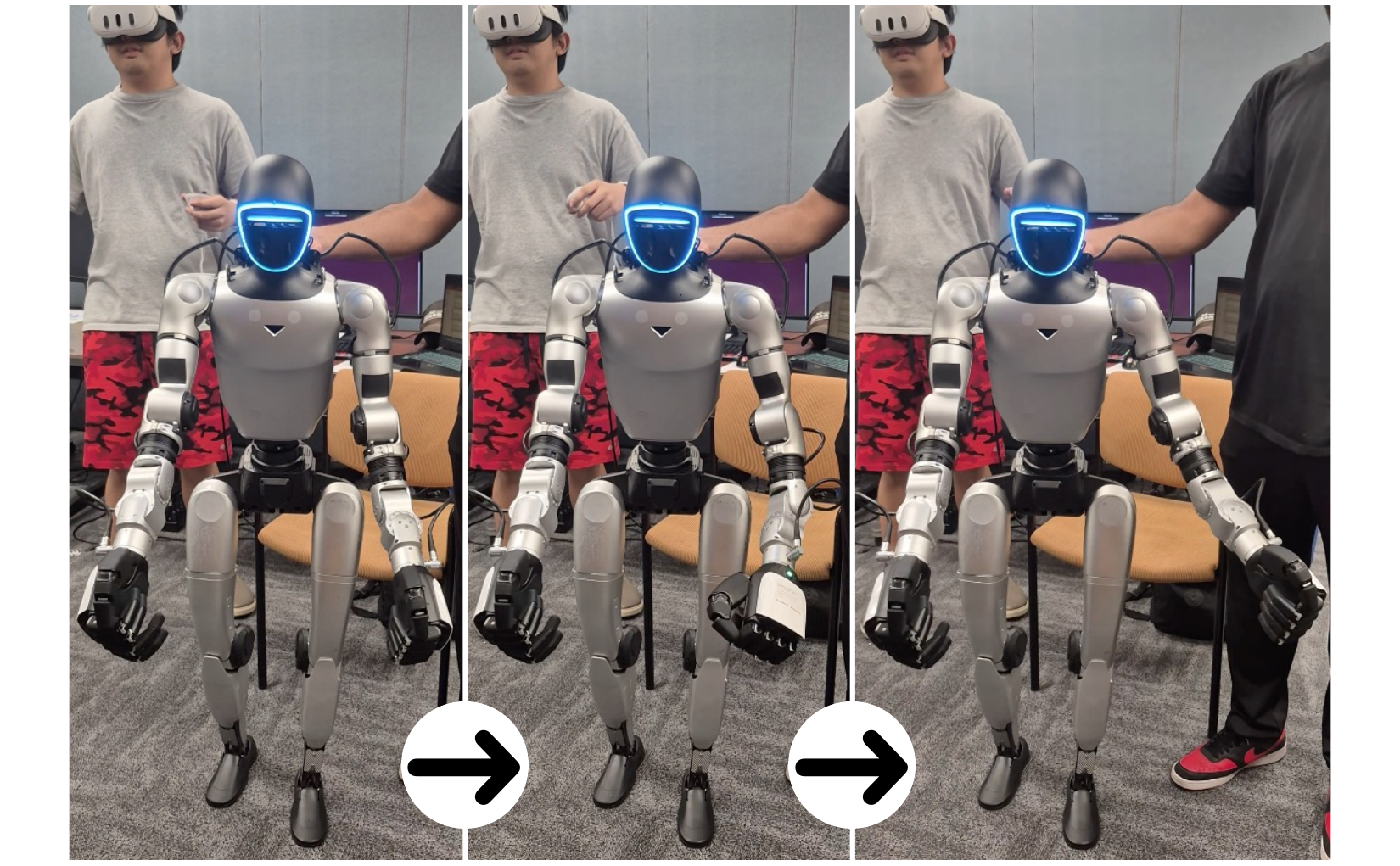}
    \caption{The hybrid control framework applied to the Unitree G1 humanoid using a Meta Quest 3 headset and controllers.}
    \label{fig:g1}
\end{figure} 

As shown in \fig{fig:precision}, post hoc comparisons indicated significantly slower input speeds during the error-critical task compared to the three other tasks (all with $p < .01$). We also observed that when in close proximity with the targets (distance between the robot end-effector and the target was less than 0.35 meters), participants either utilized more free-form arm control or more constrained torso/base control to reach and manipulate them. The correlation analysis further revealed that using more constrained torso/base control resulted in significantly fewer errors ($p < .05$) (\fig{fig:precision}, right). 

\textit{Modeling Human Behavior ---} We showed that separating control degrees of freedom when approaching targets can increase task accuracy, and participants greatly slowed down their hand movements when performing tasks requiring fine manipulation. This highlights the importance of modeling human control behavior to enable more accurate indications of when and what type of assistance should be provided. For example, control speed mapping could be dynamically adjusted to stabilize the end-effector based on user input speed~\cite{krishnan2022design}, or irrelevant control degrees could be deactivated based on the user's primary control direction~\cite{pruks2022method}.

\subsection{Task Workload}

We calculated control discontinuation by adding the duration when the user is not activating any robot actions. The results show that the workload-adaptive task resulted in a significantly higher percentage of control discontinuation (all with $p < .001$) and higher mental demand (all with $p < .01$) compared to the other three tasks (\fig{fig:workload}).

\textit{Modeling User State ---} As we showed, the impact of mental effort is non-trivial for smooth robot control. It is crucial to model user state in terms of workload during human-robot interaction, as this can inform the design of adaptive robot autonomy tailored to different task contexts, human capabilities, workload, and preferences. For example, to minimize mental workload, graphical user interfaces should present and emphasize only the information critical~\cite{szafir2021connecting} to the current state of the task (\eg distance to obstacles, success rate of grasping). Such real-time adaptation ensures the system can promptly respond to changing conditions and operator needs, thereby enhancing overall effectiveness. 

\section{Conclusion}\label{sec:conclusion}

We developed a VR-based hybrid control framework that enables flexible whole-body teleoperation of the TIAGo mobile manipulator. As shown in \fig{fig:g1}, the framework can also be applied to other robot platforms (e.g., the Unitree G1 humanoid) and input interfaces (e.g., Meta Quest 3). Our user study provided insights into users' control behaviors and strategies, including how they adapt to different task constraints and objectives. These findings highlight the importance of intuitive control interfaces and adaptive assistance in improving complex teleoperation.

\textit{Limitations ---} While our hybrid control effectively realize complex teleoperation, several aspects could be improved and might impact the control strategy. 
For example, we did not incorporate haptic feedback in our control system, even though tactile sensations have been proven to enhance intuitive interactions with remote environments~\cite{culbertson2018haptics}. 
Additionally, in this work, perception was limited and only based on a single camera embedded in the robot's head; future work may explore vision reconstruction~\cite{wonsick2021telemanipulation} to create an immersive virtual environment representing the 3D workspace.

\section{Acknowledgments}

\noindent
This work was supported by Malone Center for Engineering in Healthcare. \textbf{Author CRediT:} Conceptualization (TL, CH); Data Curation (TL, JC); Formal Analysis (TL); Investigation (TL); Methodology (all); Software (JC); Supervision (CH); Visualization (TL); Original Draft (TL); Review \& Editing (CH).
\textbf{AI Statement:} Text edited with an LLM and verified for accuracy by the authors.

\bibliographystyle{IEEEtran}
\bibliography{references}

\end{document}